%% file: neurips.tex
\documentclass{article}

\usepackage[final]{neurips_2023}

\usepackage[utf8]{inputenc} 
\usepackage[T1]{fontenc}    
\usepackage{hyperref}       
\usepackage{url}            
\usepackage{booktabs}       
\usepackage{amsfonts}       
\usepackage{nicefrac}       
\usepackage{microtype}      
\usepackage{xcolor}         
\usepackage[flushleft]{threeparttable}
\usepackage[ruled,vlined]{algorithm2e}
\usepackage{graphicx}

\title{FDAPT: Federated Domain-Adaptive Pre-Training for Language Models}

\author{%
Lekang Jiang$^{\dagger}$, Filip Svoboda$^{\dagger}$, Nicholas D.~Lane$^{\dagger \diamond}$  
\vspace{+0.1cm} \\
$^\dagger$University of Cambridge\hspace{+0.75cm}$^\diamond$Flower Labs
}

\begin{document}

\maketitle

\begin{abstract}
   Foundation models (FMs) have shown prominent success in a wide range of tasks \citep{bommasani2021opportunities}. Their applicability to specific domain-task pairings relies on the availability of, both, high-quality data and significant computational resources. These challenges are not new to the field and, indeed, Federated Learning (FL) has been shown to be a promising solution in similar setups \citep{yu2023federated, zhuang2023foundation}. This paper tackles the specific case of Domain-Adaptive Pre-Training (DAPT), a key step in the application of FMs. We conduct the first comprehensive empirical study to evaluate the performance of Federated Domain-Adaptive Pre-Training (FDAPT). We demonstrate that FDAPT can maintain competitive downstream task performance to the centralized baseline in both IID and non-IID situations. Finally, we propose a novel algorithm, Frozen Federated Domain-Adaptive Pre-Training (FFDAPT). FFDAPT improves the computational efficiency by 12.1\% on average and exhibits similar downstream task performance to vanilla FDAPT, with general performance fluctuations remaining less than 1\%. 
\end{abstract}

\section{Introduction}
Foundation models (FMs) are trained on broad data and can be adapted to different downstream tasks \citep{bommasani2021opportunities}. Recently, these models, such as GPT-4 \citep{2023arXiv230308774O} and LLaMA \citep{touvron2023llama}, have demonstrated remarkable capabilities across numerous tasks and domains, especially in the field of natural language processing (NLP) \citep{zhou2023comprehensive}. Researchers can create high-quality models by tuning FMs to specific tasks instead of building bespoke models from scratch. This approach faces challenges due to limited availability of public data \citep{villalobos2022will} and high demand of computation power \citep{bommasani2021opportunities}. To solve these problems, researchers incorporate Federated Learning (FL) \citep{mcmahan2017communication} into FMs \citep{yu2023federated, zhuang2023foundation}. FL is a decentralized approach, which allows multiple clients to collaboratively train a joint model without exchanging raw data from each client. By adopting FL when training FMs, we can leverage more distributed data and computation resources across different sources to improve model performance while preserving data privacy. 


In this research, we focus on the combination of FL and Domain-Adaptive Pre-Training (DAPT) \citep{gururangan2020don}. The aim of DAPT is to adapt original FMs to new domains by continuing the pre-training task in the target domain. Consequently, the domain-specific FMs can achieve higher performance than original models, such as Clinical BERT \citep{alsentzer2019publicly} and BioBERT \citep{lee2020biobert} in clinical and biomedical domains respectively. Combining DAPT with FL can offer the following advantages: 1) \textbf{Privacy guarantee.} Sensitive data can be utilized without direct sharing with other clients or the server to protect data privacy. 2) \textbf{Enhanced performance.} More enhanced and powerful domain-specific Pre-trained Language Models (PLMs) can be developed by training on extensive private and distributed data. 3) \textbf{Cost saving.} Only raw text data are needed, saving the substantial costs and expenses of data labelling. 4) \textbf{Wide applicability.} Domain-specific PLMs can be fine-tuned on any downstream tasks within the domain to improve performance.

Despite the potential advantages, almost no studies have investigated Federated Domain-Adaptive Pre-Training (FDAPT). The previous work conducted simple experiments with a fixed number of clients and limited experimental situations, showing that pre-training in federated manners is applicable with some decline in accuracy \citep{liu2020federated}. This initial study leaves substantial research gaps in thoroughly assessing the performance and developing novel algorithms to improve the results.


The main contributions of this research are: 1) We derive the formal definition of non-IIDness (non-independent and identically distributed) in the context of FDAPT. 2) We design a comprehensive empirical study to evaluate the performance of standard FDAPT, and conduct extensive experiments to obtain valuable results. 3) We propose Frozen Federated Domain-Adaptive Pre-Training (FFDAPT), a straightforward but effective algorithm, which improves the computation efficiency by 12.1\% and remains similar performance to the vanilla FDAPT. 4) Through a critical evaluation of our work, we identify promising future research directions for this new research area.

\section{Related Work}
\subsection{Domain-adaptive Pre-training}

Language models are often pre-trained on heterogeneous corpora, such as Wikipedia, to capture general knowledge of languages. However, these models are task-agnostic and lack domain awareness. For example, linguistic characteristics between general texts and clinical narratives are different \citep{alsentzer2019publicly}. Clinical narratives contain specialized medical terminology and abbreviations that are not commonly found in general text. These differences downgrade the performance of PLMs on tasks in specific domains. Hence, researchers propose the DAPT methods to adapt language models to target domains and tasks \citep{guo2022domain}. 

There are two types of DAPT approaches for PLMs. The first method is to continue the pre-training task on abundant unlabeled domain-specific texts without changing the loss functions and model structures. Studies demonstrated the effectiveness of this straightforward approach in various domains. For example, an empirical study \citep{gururangan2020don} conducted experiments in 4 domains, including biomedical, computer science publications, news, and reviews. It illustrated that DAPT can result in performance increases under both high- and low-resource settings. In addition, domain-specific models trained by DAPT, such as Clinical BERT \citep{alsentzer2019publicly} and BioBERT \citep{lee2020biobert}, can significantly outperform the original model on tasks in those domains. The alternative DAPT method is to add domain-distinguish pre-training tasks to the original training objective. This approach is more complicated but can be more effective for domain adaptation. For example, studies showed that incorporating adversarial domain discrimination into DAPT can enhance domain-invariant features \citep{du2020adversarial}, which further improves the model performance.

\subsection{Federated Pre-training}

Researchers trained Word2Vec \citep{mikolov2013efficient} from scratch using FL on Wikipedia and compared the performance with centralized Word2Vec \citep{ bernal2021federated}. The results showed that combining Word2Vec and FL can achieve enhanced word representations, with better similarity, analogy, and categorization outcomes. Both model quality and convergence time in FL settings are comparable to centralized training. 

FedBERT \citep{tian2022fedbert} incorporated Split Learning \citep{gupta2018distributed} with FL to resolve computational constraints in cross-device settings. It showed that the transformer layer is computationally costly to train on edge devices. Thus, it adopted Split Learning to update the transformer layer on the powerful server and conducted other layers on edge devices. The results demonstrated that Federated Split Learning can reach same performance compared to standard FL. 

While the above work focused on pre-training in the general domain, a proof-of-concept study conducted experiments in the DAPT process \citep{liu2020federated}. It continued pre-training BERT \citep{devlin2018bert} using FL with 5 clients on the clinical dataset and tested on 2 downstream tasks. Notably, different situations of data distributions are not considered in the experiments. The outcome indicated that pre-training and fine-tuning of BERT are applicable to FL settings with some decline in accuracy. 

\section{Methodology}
\subsection{Experimental Setup}

\textbf{Training and Evaluation.} In the FDAPT process, we initialize each client's model with the weights of a PLM and continue training on domain-specific datasets using the same pre-training tasks under various federated settings. Specifically, we adopt the Federated Averaging (FedAvg) \citep{mcmahan2017communication} algorithm for our experiments. To evaluate domain-specific PLMs, we fine-tune them on different downstream tasks within the domain and compare their performance. Details of FDAPT system design are illustrated in Appendix \ref{system} and evaluation metrics are introduced in Appendix \ref{evaluationmetric}. 

\textbf{Model Architecture and Framework.} As the purpose of this empirical study is to demonstrate the effectiveness of FDAPT in comprehensive experimental settings we chose the DistilBERT \citep{Sanh2019DistilBERTAD} model due to its efficiency and representativeness. Although GPT series models \citep{2023arXiv230308774O} have achieved incredible performance recently, their enormous model sizes would limit the scale of our experiments substantially. Furthermore, we used Flower \citep{beutel2020flower}, a user-friendly FL framework, for experimental simulations. Flower provides a flexible and accessible environment to easily customize FL configurations, simplifying the implementation and execution of FL experiments. 

\textbf{Datasets. } We apply FDAPT to the PubMed dataset and 9 domain-specific datasets listed in Table \ref{t:datadown}. The PubMed dataset contains abstracts and full-text research articles in the biomedical domain. The PubMed dataset adopted in our experiments \citep{cohan2018discourse} is smaller than the version used for BioBERT \citep{lee2020biobert}, which minimizes the experimental costs while maintaining pronounced results. The domain-specific PLMs are evaluated on 9 publicly available downstream tasks in the biomedical domain, including 6 Named Entity Recognition (NER), 2 Relation Extraction (RE) and 1 Question Answering (QA) datasets, as shown in Table \ref{t:datadown}.

\input{tabs/datasets}

\subsection{Non-IIDness in Federated Pre-training}

Existing research in federated pre-training methods only focused on IID situations and ignored the non-IID issue, which commonly exists in practical FL applications and can cause performance drops. Non-IIDness in supervised learning is usually related to the distribution of data labels and features \citep{kairouz2021advances}. However, datasets used for pre-training only contain raw texts, which do not have pre-defined labels or features. Therefore, we define three types of non-IIDness in the federated pre-training process, including quantity skew, sentence length distribution skew and vocabulary distribution skew.  

\textbf{Quantity skew} refers to the imbalance in the number of training data across different clients. In the context of federated pre-training, we define it as the imbalance of the number of raw texts among all clients. Additionally, sentence length and the number of vocabulary are important features of text data. Therefore, we argue that federated pre-training with imbalanced average sentence length or the number of vocabularies may affect the final results. We define the \textbf{sentence length distribution skew} as the imbalance of the average sentence length across multiple clients, and \textbf{vocabulary distribution skew} as the imbalance of the number of unique words across all clients. When creating these skews, the aim is to maximize a single metric discrepancy among all clients, while keeping other metrics almost the same. Detailed description of non-IIDness in federated pre-training is included in Appendix \ref{noniidness}. The data distribution for experiments is reported in Appendix \ref{datadistribution}. 

\subsection{Frozen Federated Domain-Adaptive Pre-Training}

The pre-training process often requires extensive computational resources. For example, it took 23 days to train the domain-specific BioBERT model on eight V100 GPUs \citep{lee2020biobert}. Hence, mitigating computation costs is important in the FDAPT process. 

Freezing specific layers or parameters of a model during fine-tuning or transfer learning aims to preserve the pre-trained knowledge encoded within those layers, while allowing other parts of the model to adapt to new tasks or domains. Studies showed that fine-tuning only a fourth of model layers can achieve 90\% of the original model quality \citep{lee2019would}. Therefore, we propose FFDAPT, a simple but effective method that incorporates freezing methods into FDAPT to improve efficiency. We illustrate the FFDAPT approach in Algorithm \ref{alg:FFDP}.

\begin{algorithm}[htbp]
	\SetKwInOut{Input}{Input}
	\SetKwInOut{Output}{Output}
	\Input{Initialized weights of $N$-layer model $\{W_{k,1}, \ldots ,W_{k,N}\}_{k=1}^K$ from $K$ clients, number of training samples for each client $\{n_{1}, \ldots ,n_{K}\}$, number of training round $T$, maximum number of frozen layers $\varepsilon$, scaling parameter $\gamma$}
	\Output{Final global weights $\{W_{1}, \ldots ,W_{N}\}$}
	$start = 1, end = 1$\tcp*{Frozen layer index}
        $n = \sum_{k=1}^{K} n_{k }$\;
	\For{$t \in \{1, 2, \ldots, T\}$}{
            \For{ $k \in \{1, 2, \ldots, K\}$}{
                $N_{k} = min \left( \varepsilon, \left \lceil \frac{n_{k}}{n} N\right \rceil \gamma \right )$\;
                $end = start + N_{k}$\;
                \uIf{$end\le N$}{
                Train$\{W_{k,1}, \ldots ,W_{k,N}\}$ with $\{W_{k,start}, \ldots ,W_{k,end}\}$ frozen\;
                }
                \Else{
                $end = end - N$\;
                Train$\{W_{k,1}, \ldots ,W_{k,N}\}$ with $\{W_{k,start}, \ldots ,W_{k,N}\}$,$\{W_{k,1}, \ldots ,W_{k,end}\}$ frozen\;
                }
                $start = end + 1$\;
                \uIf{$start > N$}{
                $start = start - N$\;
                }
            }
            $\{W_{1}, \ldots ,W_{N}\} \gets \sum_{k=1}^{K} \frac{n_k}{n} \{  W_{k,1}, \ldots ,W_{k,N}\}$\;
	}
        \Return $\{W_{1}, \ldots ,W_{N}\}$
        
	\caption{Frozen Federated Domain-Adaptive Pre-Training (FFDAPT)}
	\label{alg:FFDP}
\end{algorithm}

For each client, a portion of the consecutive layers is frozen, and the number of frozen layers is determined by the size of the dataset. We set a scaling hyper-parameter $\gamma$ to control the actual number of frozen layers and improve flexibility. Since freezing all layers are meaningless during training, $\varepsilon$ is set to be the maximum number of frozen layers. Through our method, each client freezes some layers during training to improve the computation efficiency.

\input{tabs/dt1}
\section{Results}

We report all the experimental results in Table \ref{t:result}. We demonstrate that FDAPT achieves competitive performance on downstream tasks against the centralized baseline in both IID and non-IID situations. Moreover, FFDAPT improves the training efficiency by approximately 12.1\% on average, while maintaining similar downstream tasks performance compared to vanilla FDAPT. 

\subsection{FDAPT}
We make the following observations regarding FDAPT. 

\textbf{The performance drops of federated models compared to the centralized baseline are acceptable.} While federated models exhibit a slight performance decrease compared to the centralized model on some tasks, all models pre-trained with FDAPT surpass the original model on all tasks. These datasets contain both word-level and sentence-level tasks, underscoring the effectiveness of FDAPT. On the other hand, the performance of federated models decreases by less than 1\% on almost all datasets compared to the centralized baseline. Hence, we argue that this level of performance reduction is tolerable to preserve data privacy. 

\textbf{Federated models occasionally outperform the centralized baseline.} Federated models trained in specific settings can outperform the centralized approach on NCBI, LINNAEUS, Species-800, GAD, EU-ADR and BioASQ 7b, including both token-level and sentence-level tasks. Notably, FDAPT with 2 clients under the IID setting increases the F1 scores on LINNAEUS by 0.8\%. Meanwhile, on the EU-ADR dataset, FDAPT with 8 clients under non-IID settings of sentence length skew, achieves the F1 score of 81.7\%, which is 1.1\% higher than the centralized approach. Furthermore, federated models outperform the centralized model on BioASQ 7b with increases up to 2.4\% in terms of strict accuracy. 

\subsection{FFDAPT}

The aim of FFDAPT is to improve computational efficiency. The improvement of training efficiency is calculated by the following equation. 
\begin{equation}
I = \frac{T-T_F}{T_F} \cdot 100\%
\label{e:train}
\end{equation}
where $T$ and $T_F$ are the round time for standard FDAPT and FFDAPT respectively. We calculate the improvement of efficiency for each experimental scenario, and report the average value of 12.1\% as the final result of improvement. 

In terms of downstream tasks performance, \textbf{FFDAPT can achieve similar outcomes compared to standard FDAPT in all situations}. The performance variations are no greater than 1\% for most of the cases. Remarkably, FFDAPT leads to higher performance than vanilla FDAPT in some specific scenarios. For example, the F1 scores on the Species-800 dataset increase by 0.8\% for models trained with 8 clients under the IID setting. Additionally, FFDAPT with 8 clients improves the F1 scores by 1.4\% on the EU-ADR dataset in non-IID settings of vocabulary distribution skew. Moreover, FFDAPT with 2 clients results in a 1.6\% performance enhancement on the QA task in the situation of vocabulary distribution skew. Although there are performance decreases in some situations, they still remain better than the original model.

\section{Discussion and Future Work}
This study makes significant contributions to FDAPT, but we acknowledge some limitations and point out promising research directions to stimulate future studies. 

\textbf{More real-world simulations.} Due to limited data and computational resources, our experiments focused on a particular set of common settings to obtain meaningful results while minimizing experimental costs. As it is possible that interesting connections could be discovered between model-domain pairs, and the process of federation, future research could expand on ours by attempting larger-scale simulations on an expanded selection of alternative model backbones and domain datasets. In addition, more complex non-IID simulations can be conducted, such as the skewness in the distribution of sentence embeddings, which is crucial in determining model quality.

\textbf{Improve computation and communication efficiency.} We demonstrate the efficacy of our FFDAPT algorithm, but further efficiency improvements may be possible. For instance, it is likely that module adapters \citep{cai2022autofednlp} could be used to mitigate computational costs, and communication-efficient algorithms, such as FedPCL \citep{tan2022federated}, could improve communication efficiency.

\textbf{Other challenges in FL. } Beyond the concerns of data privacy, addressed in this paper, Federated Learning can be used to address other concerns of distributed training. For example, it can address system heterogeneity, develop specific federated strategies and integrate privacy-enhancing approaches. All of these are likely to be fruitful extensions to this work.

\textbf{Domain-related challenges.} This research is based on the biomedical domain and suggests that FDAPT can have similar performance compared to centralized pre-training. In contrast, a previous study \citep{liu2020federated} showed that FDAPT is worse than centralized pre-training in the clinical domain. Our results are based on 9 diverse downstream datasets, in contrast to the alternative assessment, which was based on 2, highly similar, tasks. The difference in results is worth investigating, as, in the most interesting case, this could point to a special case identified by \citep{liu2020federated}, in which the specific task they chose happens to be harder to federate.

\section{Conclusion}

In this paper, we presented the first comprehensive empirical investigation of FDAPT for NLP tasks in cross-silo settings. We trained the DistilBERT model on biomedical corpora using FDAPT, and tested the model performance on 9 different task-specific datasets, including both token-level and sentence-level tasks. We formalized three types of non-IIDness in the context of FDAPT, and simulated both IID and non-IID data distributions. The results show that FDAPT can retain competitive downstream task performance to the centralized baseline across all tested scenarios. The performance drops of FDAPT are less than 1\%, most of the time, and it can occasionally outperform the centralized approach. In all experimental situations, models trained by FDAPT surpass the original, non-adapted, DistilBERT model, demonstrating the effectiveness of FDAPT. 

Furthermore, we propose our own efficiency-improving algorithm, the FFDAPT. It can increase training speed by approximately 12.1\%, on average. Additionally, FFDAPT exhibits similar downstream task performance to vanilla FDAPT, with general performance fluctuations remaining less than 1\%. In specific circumstances, FFDAPT can improve the performance by up to 1.6\%. 

Finally, we point out promising future research directions for this new research area. It is crucial to continue investigating this field, in order to advance our understanding of how Federated Learning can aid in domain adaptation, in particular, and the applicability of Foundation Models, in general.

\bibliographystyle{unsrtnat}
\bibliography{reference}


\appendix

\section{FDAPT System Design}
\label{system}
\begin{figure}[!htbp]
    \centering
    \includegraphics[width=.99\textwidth]{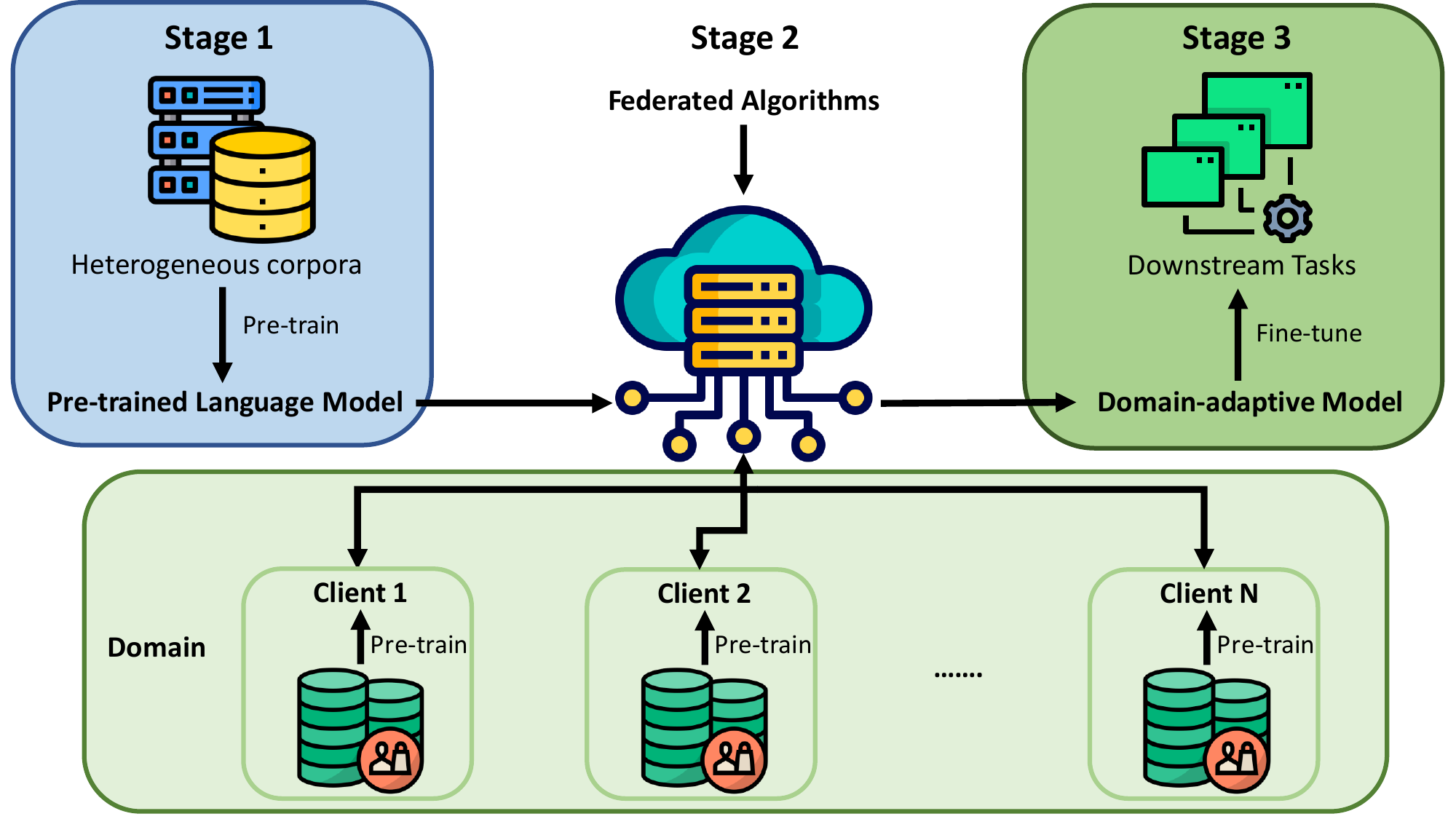}
    \caption[Overview of Federated Domain-Adaptive Pre-Training]{Overview of Federated Domain-Adaptive Pre-Training (FDAPT). Stage 1: Pre-train on heterogeneous corpora, e.g., Wikipedia. Stage 2: Apply FL in the Domain-Adaptive Pre-Training process, e.g., in the biomedical domain. Stage 3: Fine-tune on downstream tasks in the domain, e.g., disease name recognition task. }
    \label{fig:fdp}
\end{figure}

We present an overview of the FDAPT method in Figure \ref{fig:fdp}. The entire process consists of three stages. Firstly, models are pre-trained on large heterogeneous corpora, such as Wikipedia, to acquire general knowledge. These models are often trained by AI research companies and are publicly available, for example, BERT \cite{devlin2018bert} and GPT-2 \cite{radford2019language}. 

In the second stage, models continue being trained in the federated settings to adapt to specific domains. Each client trains the model on their local dataset and communicates with the server for aggregation. After a specified number of training rounds, the domain-adaptive PLMs are obtained. During this process, local datasets stored in each client are not shared to preserve data privacy, and models are aggregated to reach an enhanced global model. A group of institutes with sensitive data in the same domain can collaborate on FDAPT and release the federated domain-specific model for public usage. 

In the final stage, these models are fine-tuned on different domain-specific downstream tasks, which can achieve better performance than the original model. Even individuals or small companies that work in the same domain can fine-tune the domain-specific model on their own tasks and benefit from FDAPT.

\section{Evaluation Metrics}
\label{evaluationmetric}

NER and RE are multi-label classification tasks, which can be evaluated by \textbf{precision (P), recall (R) and F1 scores (F1)}. 

\begin{equation}
    P=\frac{TP}{TP+FP} 
\end{equation}

\begin{equation}
    R=\frac{TP}{TP+FN} 
\end{equation}

\begin{equation}
    F1=\frac{2PR}{P+R} 
\end{equation}
where TP (true positives) refers to the number of samples that are predicted positive and actually positive, FP (false positives) is the number of samples that are predicted positive but actually negative, and FN (false negatives) is the number of samples that are predicted negative but actually positive.

For each question in the factoid QA task, models can return a list of answers, ordered by decreasing confidence. We evaluate the model by strict accuracy (S), lenient accuracy (L) and mean reciprocal rank (M). 

\textbf{Strict accuracy} measures the proportion of questions, for which the first returned answer by the model contains the golden response. Assuming the golden answers are $[g_1, g_2, \ldots, g_n]$, and the predicted answers for question $k$ are $[a_{k,1}, a_{k,2}, \ldots, a_{k,m}]$, the strict accuracy can be calculated using Equation \ref{e:sa}. 

\begin{equation}
S = \frac{\sum_{k=1}^{n} (g_k \in a_{k,1})}{n}
\label{e:sa}
\end{equation}

\textbf{Lenient accuracy} is the proportion of factoid questions that have been answered correctly in the lenient sense. The results are seen as correct if one of the returned answers meets the golden standard. The formula is shown in Equation \ref{lenient}.

\begin{equation}
L = \frac{\sum_{k=1}^{n} (g_k \in [a_{k,1}, a_{k,2}, \ldots, a_{k,m}])}{n}
\label{lenient}
\end{equation}

\textbf{Mean reciprocal rank} is defined in Equation \ref{e:mrr}. It ranges from 0 to 1, and higher values indicate better performance. This evaluation metric encourages the model to return accurate answers with higher confidence. 

\begin{equation}
M = \frac{1}{n}\sum_{k=1}^{n} \frac{1}{r_k}
\label{e:mrr}
\end{equation}
where $r_k$ is the position of the first answer returned by the model that includes the golden response for question $k$

\section{Non-IIDness in Federated Pre-training}
\label{noniidness}

\textbf{Quantity skew} refers to the imbalance in the number of training data across different clients. In the context of federated pre-training, we define it as the imbalance of the number of raw texts among all clients. In our experiments, the data distribution for each client $i \in \{1, 2, \ldots, k\}$ in the situation of quantity skew is calculated using Equation \ref{e:qs}, where $Q$ is the total number of training data. 

\begin{equation}
   D_Q =  \left \{ Q_{i} \mid Q_{i} = \frac{i}{\sum_{j=1}^{k}j} Q, i \in \left \{ 1, 2, \ldots, k \right \}  \right \} 
    \label{e:qs}
\end{equation}

Sentence length and the number of vocabulary are important features of text data. Therefore, we argue that federated pre-training with imbalanced average sentence length or the number of vocabularies may affect the final results. 

We define the \textbf{sentence length distribution skew} as the imbalance of the average sentence length across multiple clients. Assuming the average sentence length of client $i \in \{1, 2, \ldots, k\}$ is $L_i$, we create the sentence length distribution skew by maximising the standard deviation of average sentence lengths of all clients. 

\begin{equation}
    D_L = \left \{ L_i \mid \max_{} \left ( \sigma \left ( L_1, L_2, \ldots, L_k \right ) \right ), i \in \left \{ 1, 2, \ldots, k \right \}     \right \} 
    \label{e:slds}
\end{equation}

Similarly, \textbf{vocabulary distribution skew} is the imbalance of the number of unique words across all clients. The data distribution of all clients in the situation of vocabulary distribution skew is illustrated in the following Equation. 

\begin{equation}
    D_V = \left \{ V_i \mid \max_{} \left ( \sigma \left ( V_1, V_2, \ldots, V_k \right ) \right ), i \in \left \{ 1, 2, \ldots, k \right \}     \right \} 
    \label{e:vds}
\end{equation}

When creating these skews, the aim is to maximise a single metric discrepancy among all client, while keeping other metrics almost the same. For example, the objective of vocabulary distribution skew is to maximise $\sigma(V_1, V_2, \ldots, V_k)$, the standard deviation of the number of vocabularies across all clients, while minimising $\sigma(Q_1, Q_2, \ldots, Q_k)$ and $\sigma(L_1, L_2, \ldots, L_k)$, the standard deviation of data quantity and average sentence length.

\section{Data Distribution}
\label{datadistribution}

We report the pre-training data distribution under different settings across 2 and 8 clients in Table \ref{t:data2}. In IID settings, the text quantity, average sentence length and average vocabulary are consistent across all clients. In each non-IID setting, only one metric is skewed, while the remaining metrics are almost uniformly distributed. For example, in quantity skew, the number of texts varies significantly among clients, but the other metrics remain stable. 

\input{tabs/data_distribution}

\section{Environmental Settings}
\label{environmental}
\subsection{Pre-training}
The federated models are trained for 15 rounds, and each client trains model for 1 epoch on the local dataset during each round, which is consistent with the previous work \cite{liu2020federated}. To make a sensible comparison, the centralised model is trained for 15 epochs. We keep the hyper-parameters unchanged when processing different experiments, using the Adam optimiser with batch size of 8 and learning rate of 5e-5. We only train each model once, because pre-training is very computationally expensive (see Section \ref{cd} for details).

\subsection{Fine-tuning}
We fine-tune the models on different downstream tasks with different hyper-parameters, base on a previous study \cite{lee2020biobert}. 

\textbf{NER.} We fine-tune models on NER tasks for 20 epochs, using Adam optimiser with batch size of 8 and learning rate of 5e-5. Models with the highest validation performance are selected and tested on the test sets to obtain the final results. 

\textbf{RE.} Since RE datasets are smaller, we fine-tune the models for 10 epochs, using Adam optimiser with batch size of 32 and learning rate of 5e-5. These datasets do not have separate test sets, so we reported the performance of 10-fold cross-validation according to previous works \cite{lee2020biobert,bhasuran2018automatic}. The dataset is divided into ten equal folds, where one fold is used for testing and the remaining nine are used for training. This process is repeated for 10 times, and the average performance is reported. 

\textbf{QA.} We train models on QA task for 10 epochs because of the limited size of dataset. Other settings are the same as we used for NER. 

Since fine-tuning is much more computationally efficient than  pre-training, all models are trained with 5 random seeds\footnote{Specifically, we use 42, 123, 3407, 43534 and 54354 for the experiments.}. The average results and standard deviations are reported to eliminate randomness.

\section{Compute Details}
\label{cd}

All the pre-training experiments are conducted on RTX 2080 Ti GPUs. The total GPU running time for pre-training is about 4640 hours. Notably, FDAPT and FFDAPT are trained with two GPUs to accelerate the training process and make a fair comparison of computational efficiency. Pre-trained models, including 1 original model, 1 centralised model and 16 federated models, are fine-tuned on downstream tasks using a single V100 GPU. Each model takes about 14 hours to finish fine-tuning on all downstream tasks. Thus, the overall GPU running time for fine-tuning is approximately $18 \times 14 = 252$ hours. 

\end{document}

%% file: tabs/datasets.tex
\begin{table}[!htbp]
\centering
\caption{ List of downstream tasks. }
\begin{tabular}{lcc}
\toprule
Datasets           & Task type & Entity type\\ \midrule
NCBI Disease \cite{dougan2014ncbi}         & NER & Disease \\
BC5CDR \cite{DBLP:journals/biodb/LiSJSWLDMWL16}              & NER  & Chemical \\
BC4CHEMD \cite{Krallinger2015TheCC}            & NER & Chemical \\
BC2GM \cite{smith2008overview}                & NER & Gene \\
LINNAEUS  \cite{Gerner2010}            & NER & Species \\
Species-800 \cite{pafilis2013species}          & NER & Species \\ \midrule
GAD \cite{Bravo2015}                  & RE & Gene-disease  \\
EU-ADR \cite{van2012eu}               & RE  & Gene-disease \\ \midrule
BioASQ 7b-factoid \cite{tsatsaronis2015overview}   & QA & N/A
\\ \bottomrule
\end{tabular}

\label{t:datadown}
\end{table}

%% file: tabs/dt1.tex
\begin{table*}[!htbp]

\resizebox{\textwidth}{!}{
\begin{threeparttable}
    \caption{Downstream tasks performance and standard deviations of the original model, the model with centralized pre-training, models with FDAPT and models with FFDAPT under IID and non-IID settings. }
    \label{t:result}
    \centering
\begin{tabular}{lcccccclcclc}
\toprule
            & \multicolumn{6}{c}{NER}                                                                                                                 &  & \multicolumn{2}{c}{RE}                      &  & QA                   \\ \cmidrule{2-7} \cmidrule{9-10} \cmidrule{12-12} 
Settings            & NCBI                 & BC5CDR               & BC4CHEMD             & BC2GM                & LINNAEUS             & Species-800          &  & GAD                  & EU-ADR                &  & BioASQ 7b            \\ \midrule
DistilBERT  & 84.6 (.4)            & 84.1 (.3)            & 86.9 (.1)            & 79.6 (.2)            & 83.5 (.8)            & 67.1 (.9)            &  & 75.8 (.4)            & 80.0 (.3)            &  & 24.9 (1.2)           \\
Centralised & 85.4 (.9)            & 86.9 (.1)            & 88.3 (.2)            & 80.8 (.4)            & 84.6 (1.6)           & 69.3 (.6)            &  & 78.0 (.1)            & 80.6 (.1)            &  & 27.0 (1.4)           \\ \midrule
\multicolumn{12}{l}{\textbf{FDAPT}} \\
\multicolumn{12}{l}{IID}                                                                                                                                                                                                         \\
2 Clients   & 85.7 (.8)            & 86.7 (.2)            & 88.0 (.1)            & 80.7 (.2)            & 85.4 (1.3)           & 68.2 (1.2)           &  & 77.7 (.3)            & 80.7 (.2)            &  & 28.0 (1.2)           \\
8 Clients   & 85.2 (.7)            & 86.1 (.2)            & 87.5 (.1)            & 80.6 (.2)            & 83.8 (.8)            & 67.8 (1.1)           &  & 77.2 (.6)            & 80.1 (.2)            &  & 28.5 (1.2)           \\

\multicolumn{12}{l}{Non-IID (Quantity Skew)}                                                                                                                                                                                     \\
2 Clients   & 85.8 (.9)            & 86.4 (.2)            & 88.2 (.1)            & 80.5 (.2)            & 85.1 (.7)            & 69.0 (.9)            &  & 78.0 (.8)            & 80.1 (.2)            &  & 28.3 (1.4)           \\
8 Clients   & 85.1 (.5)            & 86.0 (.2)            & 87.6 (.1)            & 80.1 (.5)            & 84.7 (1.4)           & 69.2 (.6)            &  & 77.5 (.2)            & 80.8 (1.2)           &  & 29.0 (.8)            \\

\multicolumn{12}{l}{Non-IID (Sentence Length Distribution Skew)}                                                                                                                                                                 \\
2 Clients   & 85.8 (.6)            & 86.5 (.2)            & 87.9 (.1)            & 80.8 (.3)            & 83.8 (1.4)           & 69.3 (.4)            &  & 78.0 (.2)            & 80.0 (.8)            &  & 27.3 (1.6)           \\
8 Clients   & 85.2 (.2)            & 86.2 (.1)            & 87.6 (.1)            & 80.5 (.5)            & 84.5 (1.2)           & 68.9 (2.0)           &  & 78.2 (.1)            & 81.7 (1.0)           &  & 29.4 (1.0)           \\

\multicolumn{12}{l}{Non-IID (Vocabulary Distribution Skew)}                                                                                                                                                                      \\
2 Clients   & 85.8 (1.2)           & 86.7 (.2)            & 88.2 (.2)            & 80.7 (.3)            & 84.6 (1.2)           & 69.4 (1.4)           &  & 77.4 (.4)            & 80.7 (.5)            &  & 26.7 (1.2)           \\
8 Clients   & 85.3 (.4)            & 86.1 (.1)            & 87.7 (.2)            & 80.7 (.2)            & 83.7 (.8)            & 68.8 (1.2)           &  & 78.0 (.1)            & 79.8 (.9)            &  & 28.8 (1.4)           \\ \midrule

\multicolumn{12}{l}{\textbf{FFDAPT}}\\
\multicolumn{12}{l}{IID}                                                                                                            \\
2 Clients   & 84.8 (.9)  & 86.3 (.2) & 87.6 (.1) & 80.3 (.1) & 85.0 (1.8) & 68.5 (1.3)  &  & 77.0 (.3)  & 79.6 (.4) &  & 28.4 (2.3) \\
8 Clients   & 85.0 (.7)  & 85.9 (.2) & 87.4 (.2) & 80.1 (.2) & 83.8 (1.2) & 68.6 (.8)   &  & 77.3 (.1)  & 80.8 (.1) &  & 26.9 (.9)  \\
\multicolumn{12}{l}{Non-IID (Quantity Skew)}                                                                                        \\
2 Clients   & 85.2 (.9)  & 86.2 (.2) & 87.7 (.1) & 80.4 (.2) & 83.7 (1.0) & 67.7 (.7)   &  & 78.5 (.1)  & 80.4 (.4) &  & 28.2 (1.2) \\
8 Clients   & 84.5 (1.0) & 86.0 (.2) & 87.4 (.1) & 80.2 (.5) & 83.5 (1.5) & 68.4 (.7)   &  & 78.2 (.2)  & 80.9 (.1) &  & 28.3 (1.3) \\
\multicolumn{12}{l}{Non-IID (Sentence Length Distribution Skew)}                                                                    \\
2 Clients   & 85.6 (.4)  & 86.4 (.1) & 87.5 (.0) & 80.3 (.3) & 84.1 (1.4) & 68.5 (.8)   &  & 78.2 (.1)  & 80.7 (.3) &  & 27.4 (1.0) \\
8 Clients   & 85.5 (.6)  & 85.9 (.3) & 87.5 (.2) & 80.6 (.3) & 83.6 (2.0) & 68.7 (1.2)  &  & 77.6 (.7)  & 80.6 (.5) &  & 27.0 (0.9) \\
\multicolumn{12}{l}{Non-IID (Vocabulary Distribution Skew)}                                                                         \\
2 Clients   & 84.8 (1.0) & 86.2 (.3) & 87.6 (.1) & 80.3 (.2) & 84.1 (1.7) & 69.7 (.8)   &  & 76.7 (.8)  & 79.0 (.3) &  & 28.3 (1.7) \\
8 Clients   & 85.5 (.6)  & 85.9 (.3) & 87.5 (.2) & 80.6 (.3) & 83.6 (2.0) & 68.7 (1.2)  &  & 77.2 (.1)  & 81.2 (.0) &  & 28.2 (.6)  \\ \bottomrule
\end{tabular}

    \begin{tablenotes}
        \item Note:  Each experiment is conducted with 5 random seeds, and the average value is reported to eliminate randomness. The evaluation metrics used in the table are F1 scores for NER and RE tasks and strict accuracy for the QA dataset. We also evaluate the precision and recall for NER and RE tasks, lenient accuracy and mean reciprocal rank for the QA dataset. The results of these metrics provide similar research outcomes. 
    \end{tablenotes}
\end{threeparttable}
}

\end{table*}

%% file: tabs/data_distribution.tex
\begin{table}[!ht]
\centering

\resizebox{\textwidth}{!}{
\begin{threeparttable}
\caption[Data distribution across 2 and 8 clients]{Data distribution across 2 and 8 clients. }

\begin{tabular}{lcccccc}
\toprule
                                  & \multicolumn{2}{c}{Quantity} & \multicolumn{2}{c}{Sentence length} & \multicolumn{2}{c}{Vocabulary} \\ \cmidrule{2-7} 
Settings                      & Average       & $\sigma$      & Average          & $\sigma$         & Average        & $\sigma$       \\ \midrule
2 Clients \\
IID                               & 60K         & 0       & 34.3            & 0            & 3.1K           & 0          \\
Quantity skew                     & 60K         & 20K        & 34.3            & 0.1            & 3.1K           & 0          \\
Sentence length distribution skew & 60K         & 0        & 34.3           & 4.7            & 3.1K           & 0          \\
Vocabulary distribution skew      & 60K         & 0        & 34.3            & 0.7            & 3.1K           & 1.3K          \\ \midrule
8 Clients \\
IID                               & 15K   & 0                  & 34.3        & 0                     & 3.1K     & 0                   \\
Quantity skew                     & 15K   & 7.6K               & 34.3        & 0.1                   & 3.1K     & 0                   \\
Sentence length distribution skew & 15K   & 0                  & 34.3        & 6.3                   & 3.1K     & 0.1K                \\
Vocabulary distribution skew      & 15K   & 0                  & 34.3        & 1.7                   & 3.1K     & 1.7K                \\ \bottomrule
\end{tabular}
\begin{tablenotes}
        \small
        \item Note: The metrics are the number of articles, average sentence length and average number of vocabularies. The average value and standard deviations $\sigma$ across 2 clients are listed. Larger $\sigma$ indicates higher value discrepancy among all clients.
    \end{tablenotes}
\end{threeparttable}

}

\label{t:data2}
\end{table}